\documentclass{article}

\usepackage{PRIMEarxiv}

\usepackage[utf8]{inputenc} 
\usepackage[T1]{fontenc}    
\usepackage{hyperref}       
\usepackage{url}            
\usepackage{booktabs}       
\usepackage{amsfonts}       
\usepackage{nicefrac}       
\usepackage{microtype}      
\usepackage{lipsum}
\usepackage{fancyhdr}       
\usepackage{graphicx}       
\graphicspath{{media/}} 
\usepackage{multicol}
\usepackage{titling}
\usepackage{algorithm}
\usepackage{algpseudocode}
\usepackage{amsmath}
\usepackage{caption}

\date{}

\title{\textbf{LEDITS: Real Image Editing with DDPM Inversion and Semantic Guidance}}

\author{
  Linoy Tsaban, Apolinário Passos \\
  HuggingFace \\
  \texttt{\{linoy, apolinario\}@huggingface.co}
}

\begin{document}
\maketitle

\newcounter{foo}
\newcounter{foo2}

\begin{figure}[htb!]
 \includegraphics[width=\linewidth]{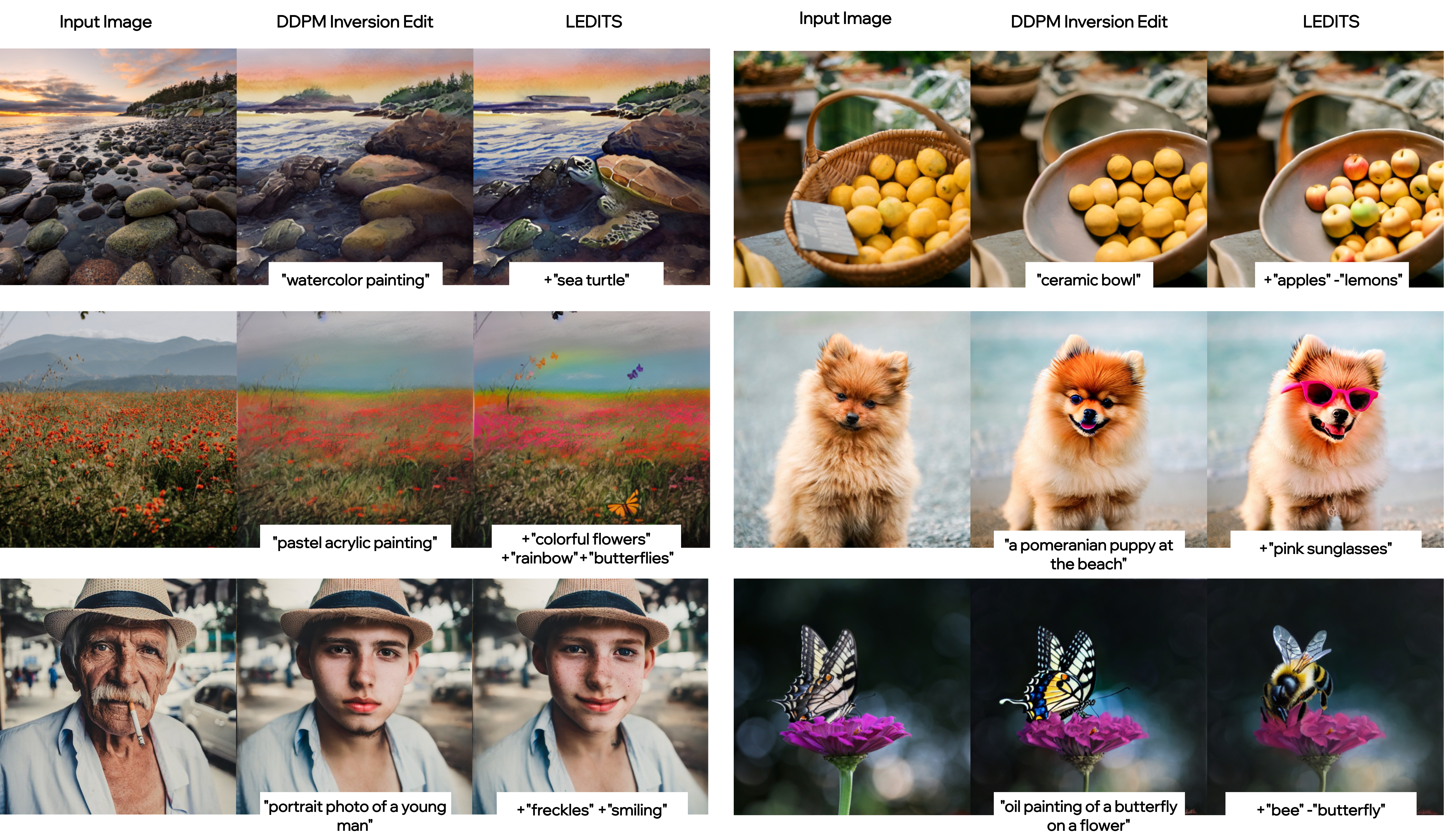}
     \caption{\textbf{LEDITS- DDPM inversion with semantic guidance for real image editing.} Real images edited purely with DDPM inversion and with both DDPM inversion and semantic guidance (LEDITS). In this combined approach we first apply DDPM Inversion on the input image, and then edit by performing the reverse diffusion process using the inverted latents and the desired target prompt, together with semantic guidance.}
 \refstepcounter{foo}\label{fig:Fig.1}
\end{figure}

\begin{abstract}
Recent large-scale text-guided diffusion models provide powerful image generation capabilities. Currently, a significant effort is given to enable the modification of these images using text only as means to offer intuitive and versatile editing. However,  editing proves to be difficult for these generative models due to the inherent nature of editing techniques, which involves preserving certain content from the original image. Conversely, in text-based models, even minor modifications to the text prompt frequently result in an entirely distinct result, making attaining one-shot generation that accurately corresponds to the user's intent exceedingly challenging. In addition, to edit a real image using these state-of-the-art tools, one must first invert the image into the pre-trained model’s domain - adding another factor affecting the edit quality, as well as latency. In this exploratory report, we propose LEDITS - a combined lightweight approach for real-image editing, incorporating the Edit Friendly DDPM inversion technique with Semantic Guidance, thus extending Semantic Guidance to real image editing, while harnessing the editing capabilities of DDPM inversion as well. This approach achieves versatile edits, both subtle and extensive as well as alterations in composition and style, while requiring no optimization nor extensions to the architecture. Code and examples are available on the project's \href{https://editing-images-project.hf.space/}{webpage}.
\end{abstract}


\section{Introduction}

The exceptional realism and diversity of image synthesis using text-guided diffusion models have garnered significant attention, leading to a surge in interest. The advent of large-scale models \cite{ho2020denoising,nichol2021glide,ramesh2022hierarchical,saharia2022photorealistic,rombach2022high, balaji2022ediffi} has sparked the imaginations of countless users, granting unprecedented creative freedom in generating images. Consequently, ongoing research endeavors have emerged, focusing on exploring ways to utilize these powerful models for image editing. Recent developments in intuitive text-based editing showcased the ability of diffusion based methods to manipulate images using text alone \cite{brack2023Sega,hertz2022prompt, couairon2022diffedit,kim2022diffusionclip, tumanyan2023plug, wallace2023edict, brooks2023instructpix2pix}. 

In a recent work by Brack et al.\cite{brack2023Sega}  the concept of semantic guidance (SEGA) for diffusion models was introduced. SEGA requires no external guidance, is calculated during the existing generation process and was demonstrated to have sophisticated image composition and editing capabilities. The concept vectors identified with SEGA were demonstrated to be robust, isolated, can be combined arbitrarily, and scale monotonically. Additional studies explored alternative methods of engaging with image generation that are rooted in semantic understanding, such as Prompt-to-Prompt \cite{hertz2022prompt}, which leverages the semantic information of the model’s cross-attention layers that associates pixels with tokens from the text prompt. While operations on the cross-attention maps enable
various changes to the generated image, SEGA does not require token-based conditioning and allows for combinations of multiple semantic changes.

Text-guided editing of a real image with state-of-the-art tools requires inverting the given image, which poses a significant challenge in leveraging them for real images. This requires finding a sequence of noise vectors that once used as input for a diffusion process, would produce the input image. The vast majority of diffusion-based editing works use the denoising diffusion implicit model (DDIM) scheme \cite{hertz2022prompt, couairon2022diffedit, tumanyan2023plug, wallace2023edict,  mokady2023null,parmar2023zero}, which is a deterministic mapping from a single noise map to a generated image. 

In the work of Huberman et al. \cite{HubermanSpiegelglas2023}, an inversion method for the  denoising diffusion probabilistic model (DDPM) scheme was proposed. They suggest a new way to compute noise maps involved in the diffusion generation process of the DDPM scheme, so that they behave differently than the ones used in regular DDPM sampling: they are correlated across timesteps and have a higher variance. Edit Friendly DDPM inversion was shown to achieve state-of-the-art results on text-based editing tasks (either by itself or in combination with other editing methods) and can generate diverse results for each input image and text, contrary to DDIM inversion-based methods. 

In this overview we aim to casually explore the combination and integration of the DDPM inversion and SEGA techniques, which we refer to as LEDITS. LEDITS consists of a simple modification to the semantically guided diffusion generation process. This modification extends the SEGA technique to real images as well as introduces a combined editing approach that makes use of the editing capabilities of both methods simultaneously, showing competitive qualitative results with state-of-the-art methods.

\section{Related Work}

\subsection{Edit friendly DDPM inversion}
A significant challenge of diffusion-based methods for image editing and manipulation is the extension to real images that requires inverting the generation process. In particular, inversion of DDPM sampling scheme \cite{ho2020denoising} posed a major challenge that was recently addressed by Huberman et al. \cite{HubermanSpiegelglas2023}. In their work, they suggest an alternative inversion, that consists of a novel way to compute the $T+1$ noise maps involved in the diffusion generation process of the DDPM scheme, so that they are better suited for editing.  \newline


In the DDPM sampling scheme, the reverse diffusion process starts from a
random noise vector $x_T \sim N(0,\mathcal{I})$ and iteratively denoises
it using 
\begin{equation} \label{eq. 1}
    x_{t-1} = \hat{\mu_t}(x_{t}) + \sigma_t z_t \hspace{1cm} t=T,...,1
\end{equation}
where ${z_t}$ are iid standard normal vectors, and
\begin{equation}\label{eq. 2}
\begin{split}
    \hat{\mu_t}(x_{t}) = \sqrt{\bar\alpha_{t-1}} (x_t -\sqrt{1-\bar\alpha_t} \epsilon_{\theta_t} )/ \sqrt{\bar\alpha} + \sqrt{1-\bar\alpha_{t-1} - \sigma^2_t}\epsilon_{\theta_t}
\end{split}
\end{equation}
where $\epsilon_{\theta_t}$ is the neural network noise estimate of $x_t$, 
and $\sigma_t = \eta \beta_t (1-\bar{\alpha}_{t-1})/(1-\bar{\alpha}_t)$ where $\beta_t$ stands for a variance schedule  and $\eta \in [0,1]$ with $\eta=1$ corresponding to the original DDPM work. The edit friendly DDPM inversion method constructs the sequence $x_1,..,x_T$ such  that structures within the image $x_0$ are more strongly “imprinted” into the noise maps $z_1,...,z_T$ that are extracted by isolating $z_t$ from eq.\ref{eq. 1}.

\subsection{Semantic Guidance}
The concept of Semantic Guidance \cite{brack2023Sega} was introduced to enhance fine grained control over the generation process of text guided diffusion models. SEGA extends principles introduced in classifier-free guidance by exclusively interacting with the concepts already present in the model’s latent space. The calculation takes place within the ongoing diffusion iteration and is designed to impact the diffusion process across multiple directions. More specifically, SEGA uses multiple textual
descriptions $e_i$, representing the given target concepts of the generated image, in addition to the text prompt p.

\section{LEDITS - DDPM Inversion X SEGA}
\begin{figure}
\centering
 \includegraphics[width=.7\linewidth]{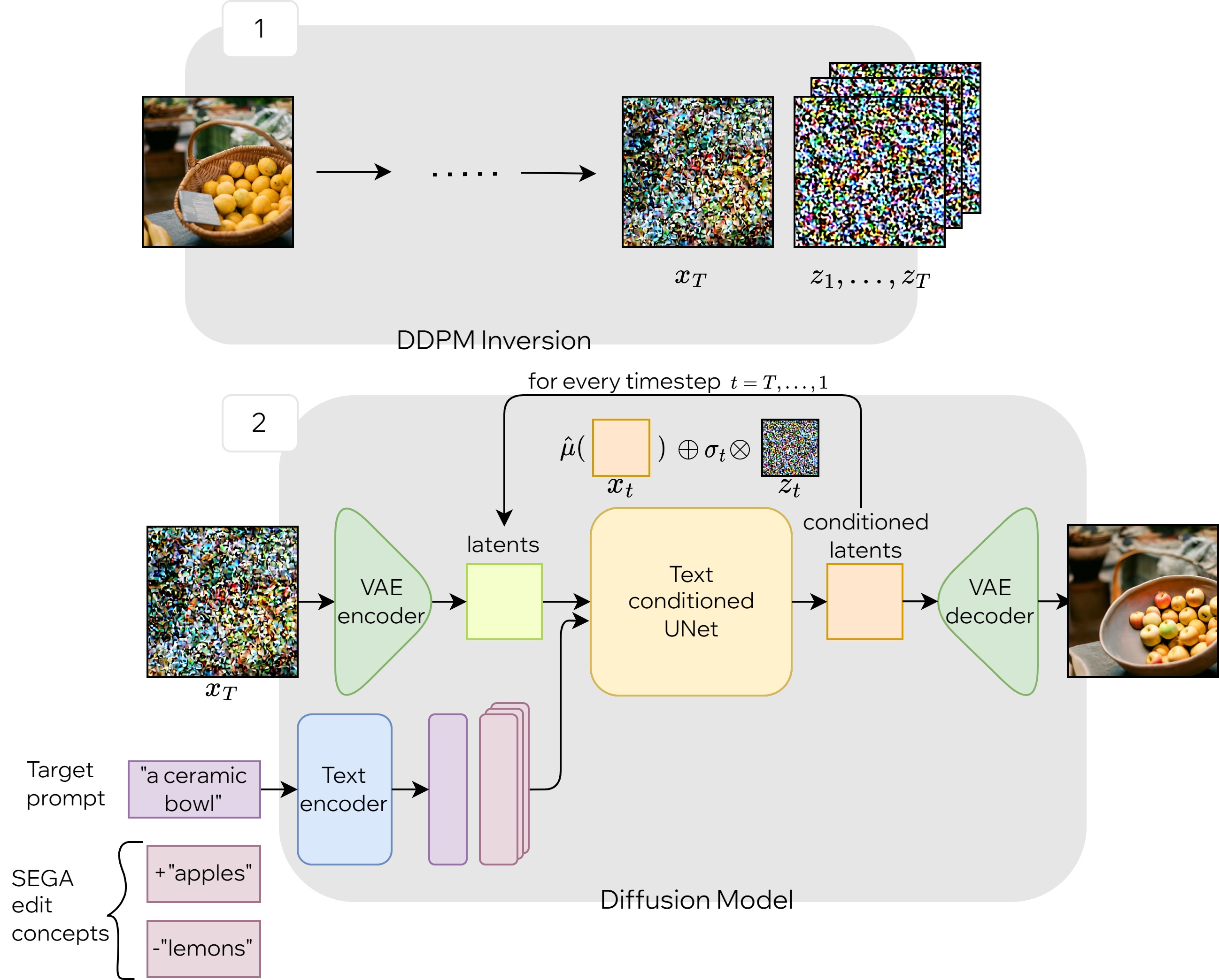}
 \caption{\textbf{LEDITS overview.} Top: inversion of the input image. We first apply DDPM inversion on the original image to obtain the inverted latents and corresponding noise maps. Bottom: We use the inverted latents to drive the reverse diffusion process with semantic guidance. In each denoising step we compute the noise estimate according to the SEGA logic and compute the updated latents according to the DDPM scheme, using pre-computed noise maps.}
 \refstepcounter{foo}\label{fig:Fig.2}
\end{figure}

We propose a straightforward integration that consists of a simple modification to the SEGA scheme of the diffusion denoising process. This modification allows the flexibility of editing with both methods while still maintaining complete control over the editing effect of each component. First, we apply DDPM inversion on the input image to estimate the latent code associated with it. To apply the editing operations, we perform the denoising loop such that for each timestep $t$, we repeat the logic used in SEGA but with the DDPM inversion scheme, using the pre-computed noise vectors. More specifically, we start the denoising process with $x_T$ computed with DDPM inversion. Let $\epsilon_{\theta_t}$ be the the diffusion model’s (DM), noise estimate with semantic guidance (following the SEGA logic) in timestep $t$. Then we update the latents according to eq.\ref{eq. 1} such that 

\begin{equation*}
   x_{t-1} = \hat{\mu_t}(x_{t}; \epsilon_{\theta_t}) + \sigma_t z_t
\end{equation*}

where $z_t$ is the corresponding noise map, obtained from the inversion process.
A pseudo-code of our method is summarized in Alg. \ref{alg:cap}. A general overview is provided in Fig. \ref{fig:Fig.2}.
\begin{algorithm}[H]
    \caption{LEDIT}\refstepcounter{foo2}\label{alg:cap}
    \renewcommand{\algorithmicrequire}{\textbf{Input:}}
    \renewcommand{\algorithmicensure}{\textbf{Output:}}
    \begin{algorithmic}[1]
    \Require Input image $I$, target prompt $p_{tar}$ and edit concepts $e_1,...,e_k$
    \Ensure Output image $\tilde{I}$
    \State Compute the inverted latent and noise maps $x_T, z_1,...,z_T$ using DDPM inversion over I;
    \State $c_{p_{tar}}, c_{e_1}, ..., c_{e_k} \gets DM.encode(p_{tar},e_1, ...,e_k)$ 
    \newline
    \For{$t=T,...,1$}{
        \State $ \epsilon_{\theta_t} = DM.predict-noise(x_t, c_{p_{tar}},  c_{e_1}, ..., c_{e_k}) $
        \State $x_{t-1} \gets \hat{\mu_t}(x_{t}; \epsilon_{\theta_t}) + \sigma_t z_t $  \Comment{update latents}\EndFor}
    
    \State $\tilde{I} \gets DM.decode(x_0)$ 
    \newline
    \Return $\tilde{I}$
    \end{algorithmic}
\end{algorithm}

\begin{figure}
 \includegraphics[width=\linewidth]{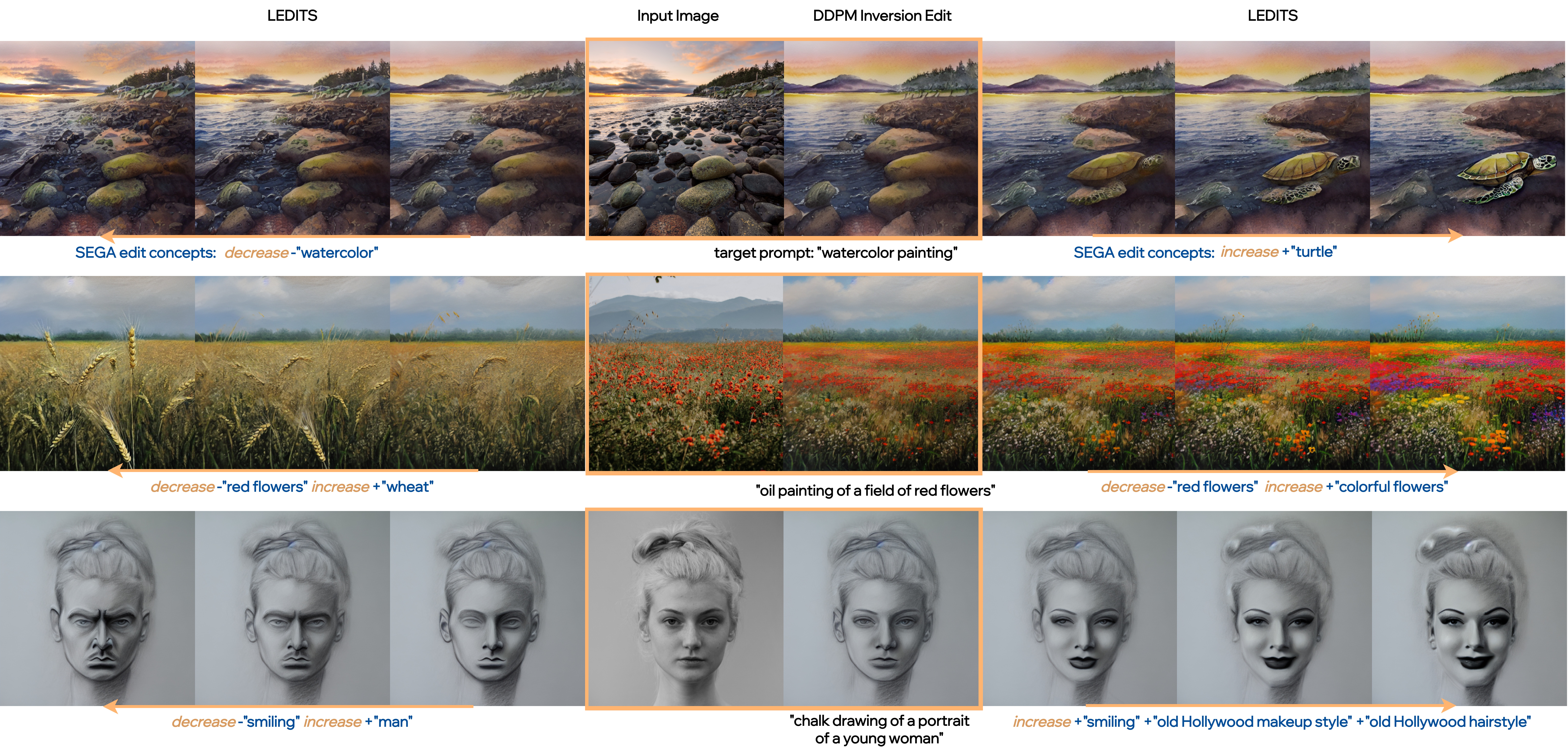}
 \caption{\textbf{Image editing with LEDITS. } LEDITS extends fine-grained control over edit operations and introduces flexibility and versatility. We show images edited purely with DDPM Inversion (forth column from the right) and images edited with LEDITS, using both methods simultaneously (three leftmost and rightmost columns) - these images were edited by using the described target prompt (in black) in addition to SEGA concepts (stated in blue). SEGA semantic vectors maintain their monotonically scaling property when used in LEDITS - the gradual effect of increasing/decreasing the strength of SEGA concepts can be observed from the third column on the right to the rightmost column, and from the third column to the left to the leftmost column.}
\end{figure}

\begin{figure}
\centering
 \includegraphics[width=.75\linewidth]{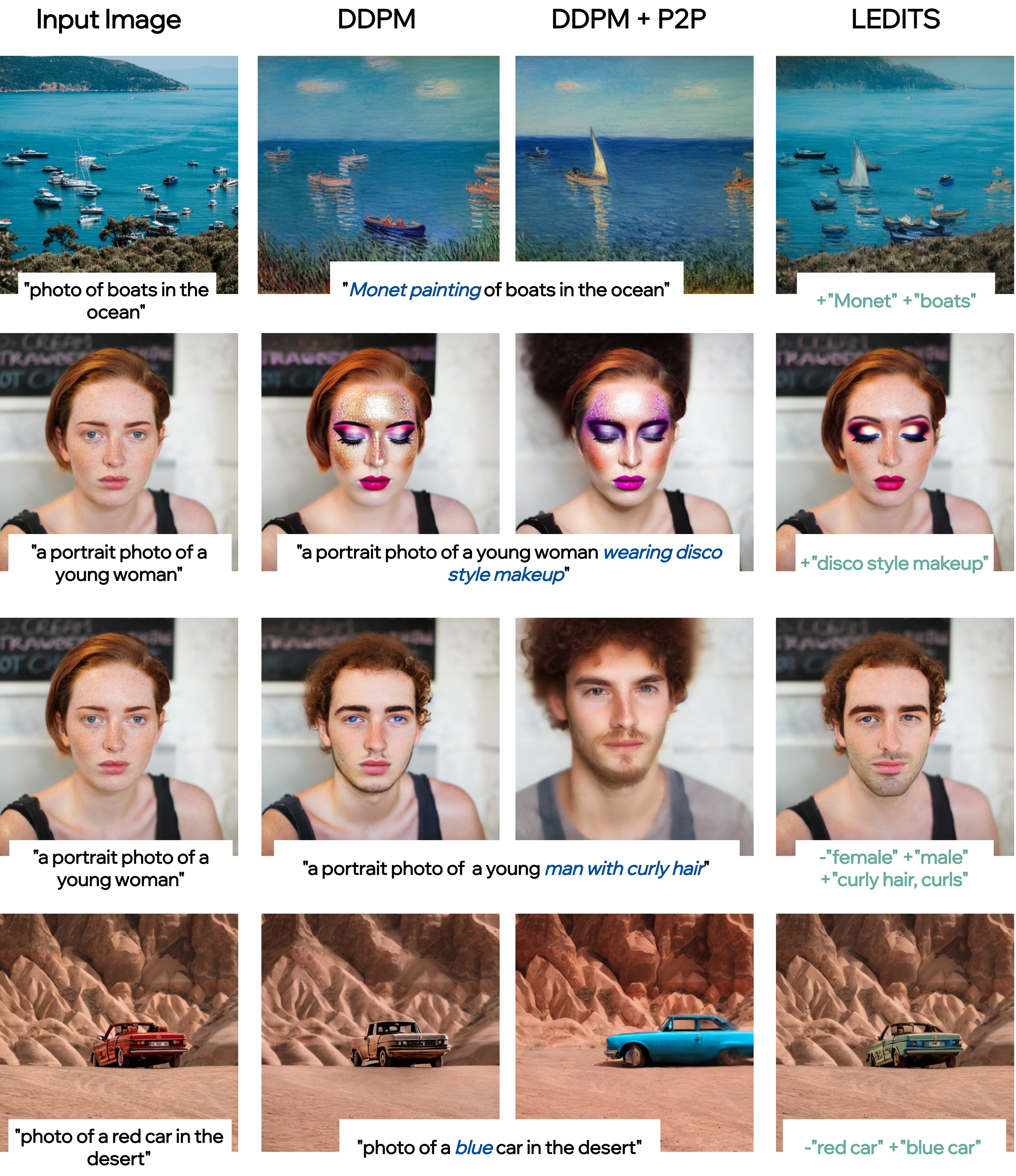}
 \caption{\textbf{Comparisons.} We show results for editing real images using pure DDPM inversions, DDPM inversion with prompt-to-prompt and LEDITS respectively. Results shown here were obtained with the first editing workflow, using DDPM purely for inversion and SEGA for editing. All images were generated using the same seed.}
\end{figure}
\section{Experiments}
\refstepcounter{foo}\label{fig:Fig.3}
\refstepcounter{foo}\label{fig:Fig.4}
We explored  two editing workflows: The first, using DDPM purely for inversion (i.e. target prompt=””), such that a perfect reconstruction of the original image is achieved and editing is done by performing semantic guidance with SEGA edit concepts. The second is performing two editing operations simultaneously by choosing a target prompt that reflects a desired output, in addition to semantic guidance with SEGA edit concepts. 

\begin{figure}
\centering
 \includegraphics[width=.7\linewidth]{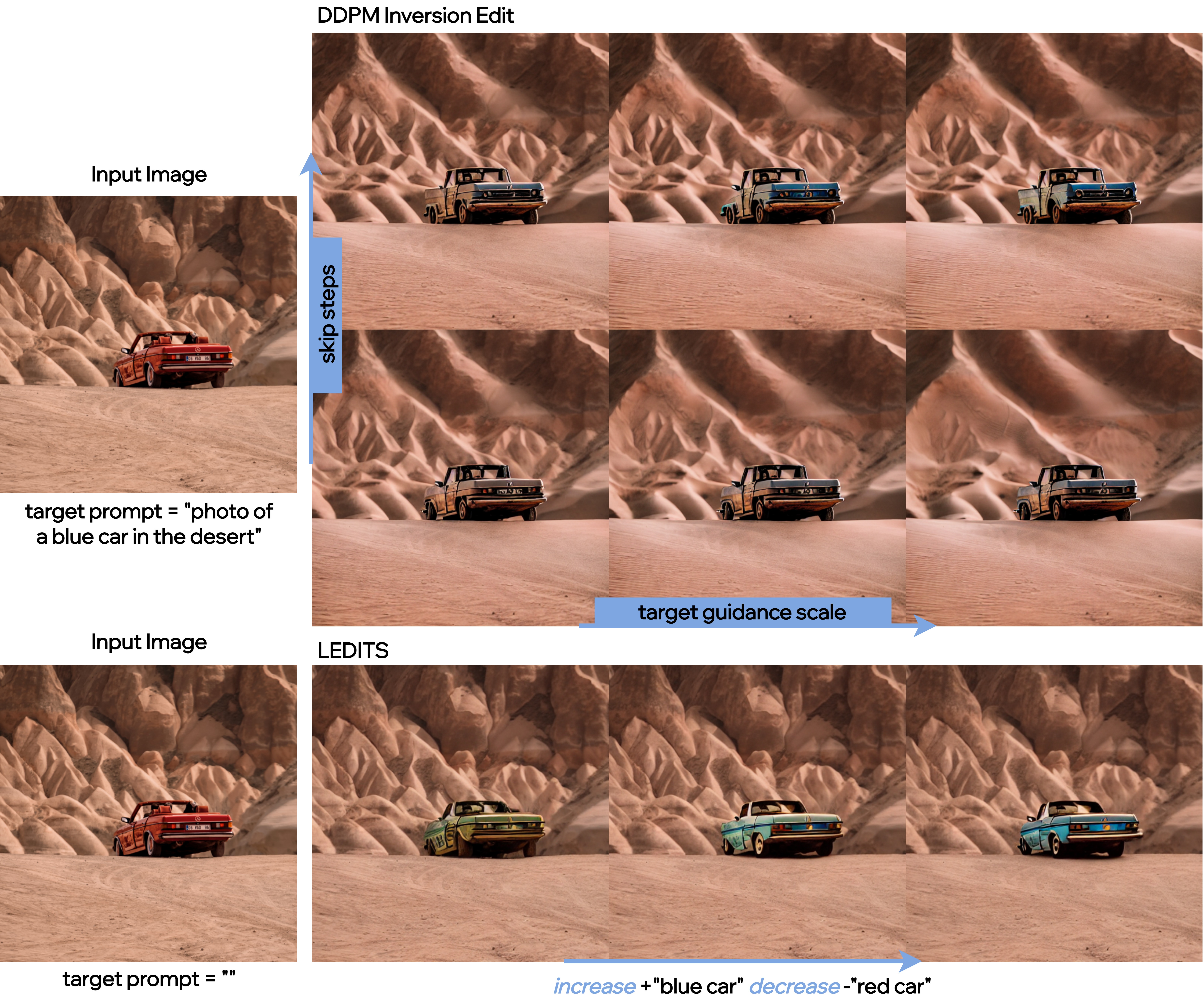}
     \caption{\textbf{Parameter effect in DDPM inversion vs. LEDITS.} We show the effect of the parameters skip steps and target guidance scale on the output image when using pure DDPM inversion (top panel) compared to the effect of the edit concepts guidance scales when using LEDITS. }
 \label{fig:Fig.5}
\end{figure}

\newpage
We observe that both approaches add diversity and versatility to the pure DDPM inversion outputs (figures \ref{fig:Fig.4} \ref{fig:Fig.5}), and extend the amount of control over edit operations. In addition, our experiments indicate that SEGA guidance vectors generally maintain their properties of robustness and monotonicity as can be seen in figures \ref{fig:Fig.1},\ref{fig:Fig.3}. Our qualitative experiments show competitive results with state-of-the-art methods and demonstrate the following properties: 
\textbf{fidelity vs. creativity} - The combined approach adds another layer of flexibility in tuning the effect of the desired edit, balancing between preserving the original image semantics and applying creative edits. \textbf{flexibility and versatility }- adding SEGA editing concepts on top of the ddpm edit (reflected in the target prompt) maintains the quality of the DDPM edit (Fig. \ref{fig:Fig.1}, \ref{fig:Fig.3}). \textbf{Complementing capabilities }- The combined control can compensate for the limitations of one approach or the other in various cases. 
In Fig. \ref{fig:Fig.5}. we explore the effect of the skip-steps and target guidance scale (the strength parameter of the classifier-free scale) parameters on the edited output, when using solely DDPM inversion for the editing operation. In comparison, we also examine the effect of SEGA concepts with increasing edit guidance scales when editing solely with SEGA (and using DDPM for inversion). We observe that the pure DDPM inversion edited outputs and pure SEGA edited outputs range differently on the scale of fidelity to the source image and compliance with the target prompt.

In addition, given the straightforward integration of the two methods, we maintain the performance advantages of the two techniques, thus making this overall approach lightweight.

\section{Conclusion}
In this report, we explored the combination of the DDPM inversion technique with semantic guidance and introduced LEDITS. We show that this efficient and lightweight approach spans a wide range of editing capabilities and extends the level of fine-grained control users have over the effect of editing operations. Our results indicate LEDITS generally maintains the individual strengths of each method, including SEGA properties such as robustness, and monotonicity. Our qualitative experiments indicate the two techniques can be used simultaneously for independent editing operations leading to more diverse outputs without harming the fidelity to the semantics of the original image and compliance with the editing prompts.

\section{Limitations}
Given the casual and exploratory nature of this report, we leave quantitative evaluations for future works and outside the scope of this work. The purpose of this report was to merely explore and suggest an intuitive editing workflow for real images, demonstrate it's qualitative abilities and potentially drive further works along this path.  

\section{Methods}
\subsection{Implementation}
The implementation of our approach builds on the Stable Diffusion and Semantic Stable Diffusion pipelines from the \href{https://github.com/huggingface/diffusers}{ HuggingFace diffusers library}. For all experiments and evaluations we used StableDiffusion-v-1-5 checkpoint. 

For the DDPM Inversion implementation, we used the official implementation at  - \href{https://github.com/inbarhub/DDPM_inversion}{https://github.com/DDPM-inversion}.

Our implementation is available on the project's \href{https://editing-images-project.hf.space/}{webpage}.

\subsection{Experiments}
All images used for our analysis were downloaded from: \href{https://www.pexels.com/}{https://www.pexels.com/}.


\begin{table}[htb!]
\centering
\begin{tabular}{||c c c c c c c||} 
 \hline
  Method & target cfg  & skip  & warm-up & threshold & edit concepts cfg &$\tau_X,\tau_a$  \\ 
 \hline\hline
 LEDITS & 15 & 36 & 1 & 0.95 & 7 & -  \\  
 DDPM & 15 & 36 & - & - & - & -  \\  
 DDPM + P2P & 9 & 12 & - & - & - & 0.6,0.2\\ [1ex] 
 \hline
\end{tabular}
 \newline \newline
\caption{\textbf{Hyper-parameters used in experiments shown in Fig. 4}. Target cfg and skip correspond to strength and $T_{skip}$ from the original DDPM inversion paper \cite{HubermanSpiegelglas2023} Edit concepts cfg corresponds to the guidance scale used for each SEGA concept individually, threshold and warm-up stand for $\lambda, \delta$ respectively from the original SEGA paper \cite{brack2023Sega}. $\tau_X,\tau_a$ are the cross- and self-attentions parameters used for P2P.}
\label{table:1}
\end{table}

In all experiments, we configured all methods to use 100 forward and backward steps. Table \ref{table:1} summarizes the huper-parameters we used for all methods to produce the results shown in Fig. \ref{fig:Fig.4}. DDPM and P2P hyper-parameters used for Fig. \ref{fig:Fig.4} were set with identical values to those used in \cite{HubermanSpiegelglas2023} for quantitative assessments.

\bibliographystyle{unsrt}  
\bibliography{references}  

\begin{thebibliography}{10}

\bibitem{ho2020denoising}
Jonathan Ho, Ajay Jain, and Pieter Abbeel.
\newblock Denoising diffusion probabilistic models.
\newblock {\em Advances in Neural Information Processing Systems},
  33:6840--6851, 2020.

\bibitem{nichol2021glide}
Alex Nichol, Prafulla Dhariwal, Aditya Ramesh, Pranav Shyam, Pamela Mishkin,
  Bob McGrew, Ilya Sutskever, and Mark Chen.
\newblock Glide: Towards photorealistic image generation and editing with
  text-guided diffusion models.
\newblock {\em arXiv preprint arXiv:2112.10741}, 2021.

\bibitem{ramesh2022hierarchical}
Aditya Ramesh, Prafulla Dhariwal, Alex Nichol, Casey Chu, and Mark Chen.
\newblock Hierarchical text-conditional image generation with clip latents.
\newblock {\em arXiv preprint arXiv:2204.06125}, 2022.

\bibitem{saharia2022photorealistic}
Chitwan Saharia, William Chan, Saurabh Saxena, Lala Li, Jay Whang, Emily~L
  Denton, Kamyar Ghasemipour, Raphael Gontijo~Lopes, Burcu Karagol~Ayan, Tim
  Salimans, et~al.
\newblock Photorealistic text-to-image diffusion models with deep language
  understanding.
\newblock {\em Advances in Neural Information Processing Systems},
  35:36479--36494, 2022.

\bibitem{rombach2022high}
Robin Rombach, Andreas Blattmann, Dominik Lorenz, Patrick Esser, and Bj{\"o}rn
  Ommer.
\newblock High-resolution image synthesis with latent diffusion models.
\newblock In {\em Proceedings of the IEEE/CVF Conference on Computer Vision and
  Pattern Recognition}, pages 10684--10695, 2022.

\bibitem{balaji2022ediffi}
Yogesh Balaji, Seungjun Nah, Xun Huang, Arash Vahdat, Jiaming Song, Karsten
  Kreis, Miika Aittala, Timo Aila, Samuli Laine, Bryan Catanzaro, et~al.
\newblock ediffi: Text-to-image diffusion models with an ensemble of expert
  denoisers.
\newblock {\em arXiv preprint arXiv:2211.01324}, 2022.

\bibitem{brack2023Sega}
Manuel Brack, Felix Friedrich, Dominik Hintersdorf, Lukas Struppek, Patrick
  Schramowski, and Kristian Kersting.
\newblock Sega: Instructing diffusion using semantic dimensions.
\newblock {\em arXiv preprint arXiv:2301.12247}, 2023.

\bibitem{hertz2022prompt}
Amir Hertz, Ron Mokady, Jay Tenenbaum, Kfir Aberman, Yael Pritch, and Daniel
  Cohen-Or.
\newblock Prompt-to-prompt image editing with cross attention control.
\newblock {\em arXiv preprint arXiv:2208.01626}, 2022.

\bibitem{couairon2022diffedit}
Guillaume Couairon, Jakob Verbeek, Holger Schwenk, and Matthieu Cord.
\newblock Diffedit: Diffusion-based semantic image editing with mask guidance.
\newblock {\em arXiv preprint arXiv:2210.11427}, 2022.

\bibitem{kim2022diffusionclip}
Gwanghyun Kim, Taesung Kwon, and Jong~Chul Ye.
\newblock Diffusionclip: Text-guided diffusion models for robust image
  manipulation.
\newblock In {\em Proceedings of the IEEE/CVF Conference on Computer Vision and
  Pattern Recognition}, pages 2426--2435, 2022.

\bibitem{tumanyan2023plug}
Narek Tumanyan, Michal Geyer, Shai Bagon, and Tali Dekel.
\newblock Plug-and-play diffusion features for text-driven image-to-image
  translation.
\newblock In {\em Proceedings of the IEEE/CVF Conference on Computer Vision and
  Pattern Recognition}, pages 1921--1930, 2023.

\bibitem{wallace2023edict}
Bram Wallace, Akash Gokul, and Nikhil Naik.
\newblock Edict: Exact diffusion inversion via coupled transformations.
\newblock In {\em Proceedings of the IEEE/CVF Conference on Computer Vision and
  Pattern Recognition}, pages 22532--22541, 2023.

\bibitem{brooks2023instructpix2pix}
Tim Brooks, Aleksander Holynski, and Alexei~A Efros.
\newblock Instructpix2pix: Learning to follow image editing instructions.
\newblock In {\em Proceedings of the IEEE/CVF Conference on Computer Vision and
  Pattern Recognition}, pages 18392--18402, 2023.

\bibitem{mokady2023null}
Ron Mokady, Amir Hertz, Kfir Aberman, Yael Pritch, and Daniel Cohen-Or.
\newblock Null-text inversion for editing real images using guided diffusion
  models.
\newblock In {\em Proceedings of the IEEE/CVF Conference on Computer Vision and
  Pattern Recognition}, pages 6038--6047, 2023.

\bibitem{parmar2023zero}
Gaurav Parmar, Krishna~Kumar Singh, Richard Zhang, Yijun Li, Jingwan Lu, and
  Jun-Yan Zhu.
\newblock Zero-shot image-to-image translation.
\newblock {\em arXiv preprint arXiv:2302.03027}, 2023.

\bibitem{HubermanSpiegelglas2023}
Inbar Huberman-Spiegelglas, Vladimir Kulikov, and Tomer Michaeli.
\newblock An edit friendly ddpm noise space: Inversion and manipulations.
\newblock {\em arXiv preprint arXiv:2304.06140}, 2023.

\end{thebibliography}

\end{document}